\documentclass{article}

\usepackage[preprint]{neurips_2020}

\usepackage[utf8]{inputenc} 
\usepackage[T1]{fontenc}    
\usepackage{hyperref}       
\usepackage{url}            
\usepackage{booktabs}       
\usepackage{amsfonts}       
\usepackage{nicefrac}       
\usepackage{microtype}      
\usepackage{xcolor}
\usepackage{amsmath}
\usepackage{graphicx}
\usepackage{caption}
\usepackage{subfigure}
\usepackage{multirow}
\usepackage{adjustbox}
\usepackage{pifont}

\title{LDTC: Lifelong deep temporal clustering for multivariate time series}

\author{%
  Zhi Wang,~~Yanni Li\thanks{Yanni~Li is the corresponding author.},~~~Pingping Zheng,~~Yiyuan Jiao \\
  School of Computer Science and Technology, Xidian University, Xi'an, China \\
  zhiwang@stu.xidian.edu.cn, yannili@mail.xidian.edu.cn, \{ppingzh, yiyuan\_jiao\}@stu.xidian.edu.cn
}

\begin{document}

\maketitle

\begin{abstract}
Clustering temporal and dynamically changing multivariate time series from real-world fields, called temporal clustering for short, has been a major challenge due to inherent complexities. Although several deep temporal clustering algorithms have demonstrated a strong advantage over traditional methods in terms of model learning and clustering results, the accuracy of the few algorithms are not satisfactory. None of the existing algorithms can continuously learn new tasks and deal with the dynamic data effectively and efficiently in the sequential tasks learning. To bridge the gap and tackle these issues, this paper proposes a novel algorithm \textbf{L}ifelong \textbf{D}eep \textbf{T}emporal \textbf{C}lustering (\textbf{LDTC}), which effectively integrates dimensionality reduction and temporal clustering into an end-to-end deep unsupervised learning framework. Using a specifically designed autoencoder and jointly optimizing for both the latent representation and clustering objective, the LDTC can achieve high-quality clustering results. Moreover, unlike any previous work, the LDTC is uniquely equipped with the fully dynamic model expansion and rehearsal-based techniques to effectively learn new tasks and to tackle the dynamic data in the sequential tasks learning without the catastrophic forgetting or degradation of the model accuracy. Experiments on seven real-world multivariate time series datasets show that the LDTC is a promising method for dealing with temporal clustering issues effectively and efficiently.
\end{abstract}

\section{Introduction}

Many application areas, such as financial trading, medical monitoring, event detection, etc., endlessly generate massive multivariate time series. Since these types of data from the real-world are rarely or sparsely labeled and have considerably dynamic changes in their important features, temporal scales, and dimensionality over time, clustering these data, namely temporal clustering for short, remains severe challenges \cite{aghabozorgi2015time,brockwell1991time,yi2017grouped}. The traditional temporal clustering algorithms generally perform the successive but two independent procedures: the dimensionality reduction and clustering. For the multivariate time series’ inherent complexity natures, the traditional algorithms would inevitably result in poor precision due to the loss of formation conversion in the two successive procedures \cite{aghabozorgi2015time}.

Inspired by a great success of deep learning in many fields in recent years, e.g., image processing, computer vision, etc., few works \cite{madiraju2018deep,franceschi2019unsupervised,yi2017grouped} started to investigate the effective and efficient deep temporal clustering methods. By integrating both the dimensionality reduction and clustering into an end-to-end deep neural network with a specific autoencoder, the few deep temporal clustering methods have demonstrated a far superior performance compared to the traditional methods. However, the accuracy and efficiency of the deep temporal clustering methods are still not perfect and need to be further improved. More importantly, none of the existing algorithms following the assumption of i.i.d. setting can capture new concepts/classes on their unsupervised sequential tasks learning \cite{chen2018lifelong}, which results in the catastrophic forgetting and unceasing degradation of the model accuracy. As stated in literature \cite{aghabozorgi2015time}, the existing temporal clustering solutions  are facing severe challenges and still have a long way to go in practice. In this paper, motivated to bridge the gap and to tackle the challenges, we propose a novel temporal clustering solution: \textbf{L}ifelong \textbf{D}eep \textbf{T}emporal \textbf{C}lustering (LDTC), which has the following unique properties: 1) With the aid of the designed new autoencoder (Dilated Causal Convolutions + Attention LSTM) with a clustering layer, it effectively integrates both the dimensionality reduction and temporal clustering  into an  end-to-end unsupervised deep learning framework resulting in clustering data effectively and efficiently, 2) on this basis, by a new hierarchical jointly optimizing of both data latent representation and clustering objective, the LDTC can achieve  high-quality clustering results comparing with the-state-of-the-art temporal clustering algorithms, and 3) by being equipped with the novel unsupervised lifelong learning mechanisms, the LDTC has gained the unique abilities to effectively deal with dynamic changes and learn new concepts/classes in the sequential tasks learning without the catastrophic forgetting and degradation of the model accuracy over its lifetime.

The remainder of this paper is organized as follows. Section 2 overviews previous works on the temporal clustering. Section 3 first outlines proposed the deep temporal clustering framework LDTC, and then discusses a novel optimizing method and lifelong learning mechanisms embedded in the LDTC. Section 4 reports experimental results, while the conclusion of this study is given in Section 5.

\section{Related Work}

In this section, we first introduce the definition of temporal clustering problem for multivariate time series, and then briefly overview the two categories of existing temporal clustering algorithms based their adopted methods.

\paragraph{Temporal clustering problem.} Given $n$ unlabeled instances/samples of multivariate time series $\{x_i \in X\}_{i=1}^{n}$ (data for short), each $x_i$ ($1 \leq  i\leq n$) from $X$ can be divided into $m$ time steps, called $m$ time dimensionality. For each time step,  $x_i$ has multiple attribute variables, i.e., features. The temporal clustering aims to cluster the data $X$ into $k$ clusters effectively and efficiently, and each cluster is represented by its cluster's centroid $\mu _j$, $j=1, ...,k$.

\paragraph{Traditional temporal clustering algorithms.} Existing traditional solutions focus mainly on dealing with one of the following two issues: an effective and efficient dimensionality reduction to filter out high frequency noise and an appropriate similarity metric to achieve the desired clustering quality \cite{aghabozorgi2015time,brockwell1991time}. However, traditional solutions have following weaknesses: 1) The dimensionality reduction is independent of the clustering criterion resulting in the potential loss of long-range temporal correlations as well as filtering out of relevant features; 2) Even with a good similarity metric, the desired clustering results are hardly obtained in the absence of the proper dimensionality reduction method due to the inherent complexity and high-dimensional nature of the data.

\paragraph{Deep temporal clustering algorithms.} In recent years, since deep learning has become a dominant approach to machine learning and has achieved great successes in many fields, a few works started to focus on supervised multivariate time series classifications with deep learning methods for achieving  better classification results  \cite{zheng2014time,pei2017multivariate,karim2019multivariate,paneri2019regularizing}. However, As multivariate time series coming from many real-world applications are rarely or sparsely labeled, unsupervised deep temporal clustering methods would be strongly preferred. Fortunately, few pioneers\cite{yi2017grouped,madiraju2018deep,franceschi2019unsupervised} have started on the exploration of the unsupervised deep temporal clustering recently. Naveen et al. \cite{madiraju2018deep} first proposed a novel deep temporal clustering algorithm DTC inspired by \cite{xie2016unsupervised}. 
The DTC integrates dimensionality reduction and temporal clustering into a deep end-to-end framework with an autoencoder embedding a novel temporal clustering. To obtain general-purpose representations for variable-length multivariate time series, based on causal dilated convolutions with a novel triplet loss employing time-based negative sampling, an unsupervised scalable representation learning method \cite{franceschi2019unsupervised} for multivariate time series  was proposed. Subin et al. \cite{yi2017grouped} designed two convolution neural networks for the temporal  clustering to obtain better results.

Although the few deep temporal clustering methods have demonstrated strong 
advantages over traditional solutions, the deep methods still suffered from the following issues more or less by our extensive experiments and 
theoretical analysis: 1) Only single-objective optimization to the latent representation learning \cite{franceschi2019unsupervised,yi2017grouped} is not enough to obtain desired clustering results, 2) the components’ configuration of their deep networks need to be further optimized for both efficient model training and high quality clustering results \cite{madiraju2018deep,franceschi2019unsupervised,yi2017grouped}, and 3) under the i.i.d. assumption, none of existing methods can capture new concepts/classes and effectively tackle the dynamic data in the unsupervised sequential tasks  learning over time effectively and efficiently.

\section{A Novel Lifelong Deep Temporal Clustering Solution LDTC}

\subsection{An overview of LDTC}

\begin{figure}[h]
    \centering
    \includegraphics[width=0.93\linewidth]{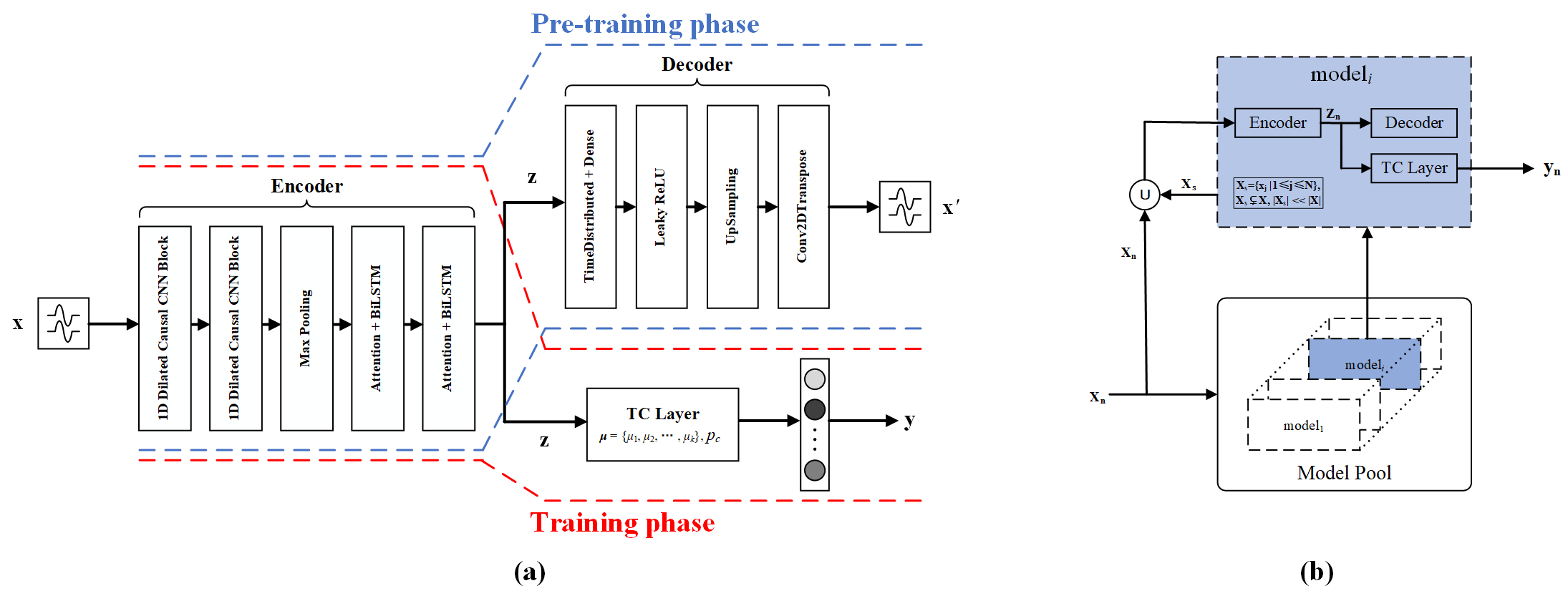}
    \vspace{-2ex}\caption{The framework of the proposed LDTC. (a) The two stages of Pre-training phase and Training phase of LDTC. (b) The schematic diagram of the lifelong learning mechanisms embedded in the LDTC, where the lifelong temporal clustering flow of the LDTC details in Sec. 3.3.}
    \label{fig1}
\end{figure}

\paragraph{A novel temporal clustering neural network.} Instead of clustering directly in the data space $X$, we would like to design a specific unsupervised deep neural network for the temporal clustering data effectively and efficiently. It first transforms the data with a non-linear mapping $f_{\theta }:X\rightarrow Z$, where $\theta$ is learnable parameters and $Z$ represents the \textit{latent features} with typically much smaller  dimensionality  than $X$ in order to avoid the "curse of dimensionality" \cite{richard1961adaptive}. Then it clusters $\{z_i \in Z\}_{i=1}^{n}$ into $k$ clusters effectively and efficiently corresponding to the data $X$. To this end, we proposed a novel \textbf{C}onvolution \textbf{T}emporal \textbf{A}uto\textbf{E}ncoder (CTAE) shown in Figure \ref{fig1}(a)

For the LDTC network, input $X$ signals are encoded into the latent space $Z$ by a convolution temporal autoencoder CTAE, which is composed of two-tiered 1D Dilated Causal CNN Blocks followed by the two-tiered Attention+BiLSTM. Each latent representation $z_i$ ($z_i \in Z$) is fed into the TC Layer (\textbf{T}emporal \textbf{C}lustering Layer) to generate its clustering assignment. ${X}'$ is the reconstructed decode corresponding to $X$ based on $Z$ by the LDTC decoder, which consists of the TimeDistributed+Dense \cite{donahue2015long}, UpSampling2D, and Conv2DTranspose \cite{zeiler2011adaptive}. 

Besides, as our ultimate goal is to achieve high-quality temporal clustering results, i.e., to meet the above third need, the representation $Z$ from CATE should have both abilities of a more compact representation and a stronger clusters’ discriminability to $X$. To this end, we introduce a new two-objective joint-optimized presentation learning strategy in a hierarchical way for effectively training $Z$, which will be discussed in Sec. 3.2. 

\paragraph{The main components for the effective latent representation learning in LDTC.} Effective latent high-level representation $Z$ of $X$, i.e., dimensionlity reduction of $X$ in latent space $Z$, is the key to high-quality temporal clustering. Generally, an effective representation $Z$ should meet three basic requirements: 1) it must extract relevant and important features from the input $X$ with the smallest reconstruction loss (see Eq. \ref{eq1}); 2) it needs to be time-efficient and memory-efficient, both for the training and testing; and 3) it helps to improve the temporal clustering quality. To meet the  first two needs with later effective implementation of the model's lifelong learning (see Sec. 3.3), we utilize a shallow deep neural network, i.e., CTAE, shown in Figure \ref{fig1}(a) to achieve these ends with two 1D exponentially Dilated Causal CNN blocks followed by two layers Attention+BiLSTM to handle multivariate time series. The reason for this design of CTAE is that the Dilated Causal CNN has been popularized in the context of sequence generation \cite{oord2016wavenet,bai2018empirical}, and that it has demonstrated unique properties in the unsupervised scalable representation learning for multivariate time series \cite{franceschi2019unsupervised}. The studies \cite{oord2016wavenet,bai2018empirical,franceschi2019unsupervised} further revealed the fact that the Dilated Causal CNN is inherently designed for sequence-modeling tasks since it not only can tackle with the issue of exploding and vanishing gradients, but also can capture key features better at a constant depth by exponentially increasing the receptive field of the network, i.e., dilation parameter ($2^i$ for $i$-th layer) compared to CNN. Moreover, the combination of attention networks \cite{chen2017recurrent,fu2019dual} and BiLSTM  \cite{hochreiter1997long,graves2005bidirectional,olah2015understanding} would further enhance the ability to capture key and long-term dependencies features of time series. 

\subsection{Model training and task referring}

It is desired to obtain an effective latent high-level representation $Z$ for the high-quality temporal clustering. However, how can we achieve this? By investigating the existing optimization methods for the issue, we found the fact that if the single objective of the reconstruction error of $X$ and $X'$ is optimized (see Figure 1(a) and Eq.\ref{eq1}), our goal will not be achieved, and that although it is theoretically feasible to optimize the reconstruction error and clustering loss at the same time, i.e., called  two-objects jointly  optimized, the training accuracy of the model will become worse and worse in practice due to the fact that they completely contradict each other. To tackle the issue, we present \textit{a new two-objective jointly optimizing strategy in a hierarchical way}.

Given an initial stochastic parameter $\theta$ of the non-linear mapping $f_{\theta}$ of CTAE, the \textbf{M}ean \textbf{S}quared \textbf{E}rror (MSE) shown in Eq. \ref{eq1} is employed to minimize the reconstruction loss error between $X$ and $X'$ for optimizing initial latent representation $Z$ corresponding to $X$, which is implemented by the Pre-training phase of the LDTC shown in Figure \ref{fig1}(a) (the top half part  with a red dotted line).

\begin{equation} \label{eq1}
L_{\textit{MSE}}=\sum _{i=1}^{n}(x_i-{x}'_i)^2 / n, \; x_i\in X, \; {x}'_i\in {X}'
\end{equation}

Using the initial optimized $Z$, the initial top $k$ centroids $\{\mu_j \}_{j=1}^{k}$ stored in TC Layer of  LDTC are obtained by performing the hierarchical clustering method \cite{pedregosa2011scikit}. After that, by iteratively refining $Z$ and $\{\mu_j \}_{j=1}^{k}$, and by jointly optimizing  the two objectives of $Z$, i.e., an effective latent representation and a stronger clusters’ discriminability of $X$,  are achieved by the training phase of LDTC shown in Figure \ref{fig1}(a) (the bottom half part with a red dotted line).

It is worth noting that we train the TC layer of the LDTC using an unsupervised algorithm that alternates between the following two steps: 1) Compute the probability shown in Eq. \ref{eq2} of assignment of the latent representation $z_i$ corresponding to input $x_i$  belonging to the cluster $j$. The closer the $z_i$ is to the centroids ${\mu_j}$, the higher the probability of $z_i$ belonging to cluster $j$; 2) Update the centroids $\{\mu_j \}_{j=1}^{k}$ by a loss function shown in Eq. \ref{eq3}, which maximizes the high confidence assignments, which will be discussed later.

When $z_i$ ($z_{i}=f_{\theta }\left ( x_{i} \right )\in Z $ corresponds to $ x_{i}\in X $ after embedding) is fed into TC Layer of LDTC shown in Figure \ref{fig1}(a) , the distance $d_{ij}$ between $z_{i}$ and $\mu _j$ is calculated using a similarity metric, e.g., Euclidean distance. With the Student’s t-distribution \cite{maaten2008visualizing} as a kernel, the probability assignment of latent $z_{i}$ belonging to $\mu _{j}$, i.e., belonging to $j^{th}$ cluster, is defined as

\begin{equation} \label{eq2}
q_{ij}= \frac{\left ( 1 + \left \| z_{i} - \mu _{j}\right \|^{2} / \alpha \right )^{-\frac{\alpha + 1}{2}}}{\sum_{{j}'=1}^{k}\left ( 1 + \left \| z_{i} - \mu _{{j}'}\right \|^{2} /\alpha \right )^{-\frac{\alpha + 1}{2}}}
\end{equation}

where $\alpha$ is the degrees of freedom of the Student’s t-distribution. In the unsupervised setting, we can set $\alpha =1$ as suggested by \cite{maaten2008visualizing}.

To refine $z_i$ and $\{\mu_j \}_{j=1}^{k}$ iteratively in the Training phase of the LDTC, the objective of minimizing the KL divergence between $q_{ij}$ and a target distribution $p_{ij}$ is formulated as

\begin{equation} \label{eq3}
L_{\textit{KLD}}=\sum_{i}^{}\sum_{j}^{}p_{ij}log\frac{p_{ij}}{q_{ij}}
\end{equation}

where $p_{ij}$ is defined as:

\begin{equation} \label{eq4}
p_{ij}=\frac{q_{ij}^{2}/f_{j}}{\sum_{{j}'=1}^{k}q_{i{j}'}^{2}/f_{{j}'}}
\end{equation}

where $f_{j}= \sum_{i}^{}q_{ij}$ are soft cluster frequencies. Note that the reason why $p_{ij}$ is chosen as the target distribution of $q_{ij}$ is that it has the following excellent characteristics   \cite{xie2016unsupervised}: 1) Strengthen predictions (i.e. improve cluster purity/accuracy), 2) put more emphasis on data points assigned with high confidence, and 3) normalize loss contribution of each centroid $\mu _j$ to prevent large clusters from distorting the latent feature space $Z$.

On this basis, we can jointly optimize latent representation $z_i$ and cluster centers $\{\mu _j\}_{j=1}^k$ stored in TC layer using Adam \cite{kingma2014adam}, a method for stochastic optimization,  with Eqs. \ref{eq5} and \ref{eq6}.

\begin{equation} \label{eq5}
\frac{\partial L_{\textit{KLD}}}{\partial z_{i}}=\frac{\alpha + 1}{\alpha }\sum_{j}^{}\left ( 1 + \frac{\left \| z_{i}-\mu _{j} \right \|^{2}}{\alpha } \right )^{-1}\times \left ( p_{ij} - q_{ij}\right )\left ( z_{i}-\mu _{j} \right )
\end{equation}

\begin{equation} \label{eq6}
\frac{\partial L_{\textit{KLD}}}{\partial \mu _{j}}=-\frac{\alpha + 1}{\alpha }\sum_{i}^{}\left ( 1 + \frac{\left \| z_{i}- \mu _{j} \right \|^{2}}{\alpha } \right )^{-1}\times \left ( p_{ij} - q_{ij}\right )\left ( z_{i}- \mu_{j} \right )
\end{equation}

The gradients $\frac{\partial L_{\textit{KLD}}}{\partial z_{i}}$ are then passed down to the CTAE of the LDTC and with standard back propagation to compute CTAE's parameters gradient $\frac{\partial L_{\textit{KLD}}}{\partial \theta}$. 

Different from the existing optimizing methods \cite{yi2017grouped,madiraju2018deep,franceschi2019unsupervised}, our LDTC first optimizes the latent  representation $z_i$, and then refines/optimizes the targets $z_i$ and ${\mu_j}$ simultaneously, so it is called the  hierarchically optimizing method, which is shown to work better than other methods in our extensive  experiments. Moreover, it is worth noting that \textit{our proposed training and optimizing strategies embedded in the Training phase of the LDTC are self-taught} \cite{raina2007self} \textit{and self-training} \cite{nigam2000analyzing} with $\{p_{ci},\{\mu _j\}_{j=1}^{k}\}$ stored in the TC Layer and $\{\textit{model}_i\}_{i=1}^{P}$ stored in the Model Pool of the LDTC.

\subsection{Lifelong temporal clustering learning of the LDTC}

So far, the model LDTC is only isolated single-task learning, i.e., it is only suitable for an independent identically distributed data, i.e., i.i.d. setting. For a  new task of the sequential tasks supervised learning,  as data of the new task do not meet the i.i.d. assumption, the model LDTC will tend to catastrophically forget existing knowledge, i.e., so-called catastrophic forgetting phenomenon. This is a long-standing challenge for machine learning,  neural network systems \cite{mccloskey1989catastrophic,goodfellow2013empirical}, and the non-stationary distributions' temporal clustering tasks especially. Although a variety of supervised lifelong learning mechanisms \cite{parisi2019continual,hu2019overcoming} have emerged to tackle the challenge, few works focused on the unsupervised scenario or multivariate time series clustering \cite{rao2019continual,smith2019unsupervised,munoz2019unsupervised,parisi2019continual}. Standing on the previous works, we present following lifelong temporal clustering learning methods embedded in LDTC for defeating the catastrophic forgetting and degeneration of model LDTC.

Note that the proposed lifelong temporal clustering learning methods are based on the Model Pool component shown in Figure \ref{fig1}(b) and the parameters, cluster centers $\{\mu_i\}_{i=1}^k$  and $p_{c_i}$ , the coverage's value of $p_{ij}$, are stored in TC layer corresponding to $\textit{model}_i$ in the component Model Pool.

\paragraph{Fully dynamic model expansion.} Let a  sequence of temporal clustering tasks be $T=(T_1, T_2,\cdots , T_M)$ corresponding to input dataset $X=(X_1, X_2,\cdots , X_M)$ and denote a  parameters set as $\theta^{(i)}=\{\theta_i, \mu_j, p_{ci}\}$ 
corresponding to the parameters of autoencoder CTAE of the LDTC, clustering centroids and the convergent value $p_{ij}$ (see Eq. \ref{eq4} ) in the Training phase on $X_i$, respectievly, where $\mu_j$ and $p_{ci}$ are stored in the TC layer of $\textit{model}_i$ in Model Pool. For each $X_i$ ($i = 1, 2, \cdots , M$), it is fed into all of the models in Model Pool in parallel, and then LDTC activates to calculate $p_{X_i}$, the average clustering accuracy of task $T_i$ through $\textit{model}_i$ testing. The following two cases need to be dealt with separately:

1) Given an empirical threshold $\delta =0.05$, the $\textit{model}_i$ (parameter set $(\theta ^{i})^{*}$) with the minimum value of $v_i=|p_{X_i}-p_{ci}| \le \delta$ is fetched from the Model Pool. And, the LDTC triggers one of the two procedures: (1) \textbf{the LDTC refines $\textit{model}_i$}, with initial parameters $\theta ^{i}_{init}$ (see Eq. \ref{eq7}), on data $X_i$ of $T_i$ with the Training phase of the LDTC when $0 \le v_i \le 0.02$ (the range of $v_i$ is a empirical one, which works well in our extensive experiments). Otherwise, (2) \textbf{the LDTC trains $\textit{model}_i$} with initial parameters $\theta ^{i}_{init}$ on data $X_i$ of $T_i$ with the two phases of Pre-train + Train of LDTC. Finally, LDTC stores the calculated parameter set $\theta^{(i)}$ corresponding to $\textit{model}_i$ and obtains the temporal clustering results of $T_i$ through the updated $\textit{model}_i$. The processing flow of the (2) is shown in Figure \ref{fig1}(b). Notice that, as the procedure (1) implements refining the $\textit{model}_i$ on data $X_i$, the performance of $\textit{model}_i$ becomes better.

2) If $v_i>\delta$, the LDTC identifies that $T_i$ is a new task. As a result, the following procedures are executed by the LDTC: (1) Instantiate a new $\textit{model}_j$; (2)  Initialize $\textit{model}_j$ with parameters $\theta ^{i}_{init}$ (see Eq. \ref{eq7}); (3) Start to train $\textit{model}_j$ on $X_i$ following the two phase of Pre-training and Training of LDTC; (4) Update the parameters $\theta^{i}_{init}$ of $\textit{model}_j$ as $\theta^{i}$; And (5) Obtain the cluster result of $T_i$ with the updated $\textit{model}_j$. Note that, as $\textit{model}_j$ is a new one, if the size of the Model Pool has not reached its capacity $P$, the well-trained model $\textit{model}_j$ will be added to the Model Pool. Otherwise, the LDTC first checks each of the habituation counter $h_j$ in $\textit{model}_j$, where  $h_j$ indicates how frequently $\textit{model}_j$ has fired based on input data. Then it pushes the model $\textit{model}_j$ with the minimum value $h_j$ into the disk, and adds the new model $\textit{model}_j$ into the pool.

\begin{equation} \label{eq7}
\theta ^{i}_{init} \leftarrow  (\theta ^{i})^{*}; \; \; \; (\theta ^{i})^{*}=\underset{i\in {1,2, \cdots M}}{\textup{argmin}}|p_{Xi}-p_{ci}|\;\;\; \textup{s.t.}\;  |p_{X_i}-p_{ci}| \le \delta
\end{equation}
where $\theta ^{i}_{init}$ is the initial parameters of $\textit{model}_i$ from Model Pool, and $(\theta ^{i})^{*}$ is the best matching model parameter values of $\textit{model}_i$ from Model Pool for $X_i$, while $p_{X_i}=(\sum_{i=1}^{|X_i|}p_{ij})/|X_i|$, the average clustering accuracy of task $T_i$ through $\textit{model}_i$. $p_{ci}$ is the coverage's value of $p_{ij}$ stored in TC layer corresponding to $\textit{model}_i$ in the component Model Pool.

As seen in the above discussion, once encountering a new task, the LDTC would automatically adapt or extend a model in Model Pool to accommodate new tasks. This lifelong machine learning method is called the fully dynamic model expansion \cite{parisi2019continual}. Notice that as the proposed the LDTC is a  shallow neural network, but with good performance, the strategy of the fully dynamic model expansion embedded in the LDTC works efficiently.

\paragraph{Resistance to forgetting and degradation via the mixture replay.} For the above procedure 1),  if $0 < v_i \le 0.02$, it means that $T_i$ follows i.i.d. setting or an approximate distribution with model $\textit{model}_i$. As an approximate distribution $T_i$ would interfere with the previous learned knowledge resulting in the issues of the catastrophic forgetting and degradation of model accuracy, LDTC should be stronger in resisting/tackling the issues. To achieve the end, we proposed a new resistant method to forgetting and degradation of LDTC model inspirited one relevant technique called Deep Generative Replay (DGR) \cite{shin2017continual}. That is, with the mixture of the original samples $X_s$ of $\textit{model}_i$ and $X_i$ of $T_i$, called mixture replay for short, the model $\textit{model}_i$ is refined by the Training phase of LDTC. Notice that the proposed rehearsal-based method \cite{parisi2019continual} can be incorporated holistically into proposed framework LDTC at a minimal cost. And our experiments demonstrated the mixture replay to be simple and effective. 

\section{Experiments}

In this study, all experiments were conducted on the server: Intel(R) Xeon(R) Gold 5115 CPU @ 2.40GHz, 97GB RAM and NVIDIA Tesla P40 graphics processor running Red Hat 4.8.5. The proposed LDTC and baseline DTC \cite{madiraju2018deep} were implemented and tested using Python 3.7, TensorFlow 1.12 and Keras 2.2.3 software. Another baseline USRL \cite{franceschi2019unsupervised} was implemented and tested using Python 3.7, PyTorch 0.4.1 software. And k-means \cite{kmeans1967} also was implemented and tested using scikit-learn 0.22.2 \cite{pedregosa2011scikit}. The hyper-parameters settings of models are shown in Table \ref{tab1}.

\renewcommand{\arraystretch}{1.5}
\begin{table}[h]
\caption{The hyper-parameters settings on evaluated models.}\label{tab1}
\vspace{10pt}
\centering
\resizebox{0.95\textwidth}{10mm}{
\begin{tabular}{cccccccc}
\hline
\textbf{Algorithm} & \textbf{Optimizer} & \textbf{Learning rate} & \textbf{Batch size} &  \textbf{Pre-train epochs}&  \textbf{Train epochs} & \textbf{Lifelong threshold} ($\mathbf{\delta}$) & $|$\textbf{Model Pool}$|$ $\mathbf{(P)}$ \\ \hline
DTC       & Adam      & 1e-3          & 64         & 10                        & 100                    & None & None \\ \hline
USRL      & Adam      & 1e-3          & 10         & None                      & 1500                   & None & None \\ \hline
LDTC      & Adam      & 1e-3          & 64         & 10                        & 100                    & 0.05 & 5 \\ \hline
\end{tabular}}
\end{table}

\subsection{Datasets and evaluated metrics}
We used seven multivariate time series datasets from real-world collected by \cite{karim2019multivariate} for evaluation, which are EEG2, NetFlow, Wafer, HAR, AREM, Uwave, and ArabicDigits, respectively. The notations $N$, $C$, $V$, and $L$ shown in Table \ref{tab2} denote the totality of samples, classes, variables and length of data, respectively.

It is worth noting that these datasets were originally used for classification and regression tasks \cite{schafer2017multivariate,lichman2013uci}. In our experiments, we mix the training and test dataset of each original dataset as the training dataset for the temporal clustering testing.

We use the standard unsupervised evaluation metrics Accuracy and Purity to evaluate the models, and we only use the labels of ground-truth categories for evaluations. The Accuracy and Purity are defined as follows, respectively.

\begin{equation} \label{eq8}
Accuracy= \frac{\sum_{i= 1}^{n} \textbf{\textit{I}} \left ( y_{i},map\left ( \hat{y}_{i} \right ) \right )}{n}; \;\;\; Purity=\frac{1}{n}\sum_{k} \underset{j}{max}  \left | C_{k} \cap L_{j}\right |
\end{equation}
where $y_{i}$ is the clustering label of $x_i$, and $\hat{y}_{i}$ is the  ground-truth categories label of $x_i$. The value of indicator function $\textbf{\textit{I}}(\cdot )$ is equal to 1, if $y_{i}= \hat{y}_{i}$, otherwise, it is equal to 0. The label of ground-truth category $C_{k}= \{ x_{i} \mid \hat{y}_{i}= k\}$ is $k$ ($k= 1,2,\cdots ,K$), and the label of cluster $L_{j} = \{ x_{i} \mid y_{i}= j\}$ is $j$ ($j= 1,2\cdots ,J$).

\subsection{Experimental results}

\textbf{Unsupervised i.i.d. temporal clustering.} We first report the evaluated results (the average of ten experimental results) of the i.i.d. temporal clustering accuracy/purity of proposed LDTC and other baselines k-means \cite{kmeans1967}, DTC \cite{madiraju2018deep} and USRL \cite{franceschi2019unsupervised} on seven real-world datasets with different training samples (N), series length (L), the number of variables (V) and categories (C). The results are shown in Table \ref{tab2}.

\begin{center}
\renewcommand\arraystretch{1.5}
\begin{table}[h]
\caption{The performance comparison of different temporal clustering algorithms on seven datasets.}\label{tab2}
\vspace{10pt}
\centering
\resizebox{0.98\textwidth}{20mm}{
\begin{tabular}{ccccccccc}
\hline
\multirow{2}{*}{\begin{tabular}[c]{@{}c@{}}\textbf{Datasets}\\ \textbf{(N, L, V, C)}\end{tabular}} & \multicolumn{2}{c}{\textbf{k-means}} & \multicolumn{2}{c}{\textbf{DTC}} & \multicolumn{2}{c}{\textbf{USRL}} & \multicolumn{2}{c}{\textbf{LDTC}} \\ \cline{2-9}& Accuracy     & Purity       & Accuracy   & Purity     & Accuracy    & Purity     & Accuracy    & Purity     \\ \hline
\multicolumn{1}{l}{EEG2 (1200, 256, 64, 2)}               & 0.570833     & 0.570833     & 0.568833   & 0.568833   & 0.584500    & 0.584500   & \textbf{0.596700}    & \textbf{0.596700}   \\ \hline
\multicolumn{1}{l}{NetFlow (1337, 994, 4, 2)}            & 0.778908     & 0.779357     & 0.808527   & 0.808527   & 0.786275    & 0.779357   & \textbf{0.825564}    & \textbf{0.825564}   \\ \hline
\multicolumn{1}{l}{Wafer (1194, 198, 6, 2)}               & 0.503853     & 0.893635     & 0.514071   & 0.893635   & 0.546901    & 0.893635   & \textbf{0.725461}    & \textbf{0.893635}   \\ \hline
\multicolumn{1}{l}{HAR (10299, 128, 9, 6)}               & 0.453636     & 0.492320     & 0.524187   & 0.560540   & 0.659773    & 0.659579   & \textbf{0.660479}    & \textbf{0.682602}   \\ \hline
\multicolumn{1}{l}{AREM (82, 480, 7, 7)}                  & 0.690244     & 0.724390     & 0.556097   & 0.580488   & 0.682927    & 0.716098   & \textbf{0.704878}    & \textbf{0.766585}   \\ \hline
\multicolumn{1}{l}{Uwave (4478, 315, 3, 8)}              & 0.699464     & 0.722242     & 0.592809   & 0.599643   & 0.767057    & 0.783154   & \textbf{0.807880}    & \textbf{0.798307}   \\ \hline
\multicolumn{1}{l}{ArabicDigits (8799, 93, 13, 10)}      & 0.504353     & 0.504353     & 0.587135   & 0.597543   & 0.543345    & 0.573577   & \textbf{0.715900}    & \textbf{0.715900}   \\ \hline
\end{tabular}}
\end{table}
\end{center}

From Table \ref{tab2}, the following conclusions can be clearly drawn: 

1) All the deep temporal clustering algorithms DTC, USRL and LDTC outperform the traditional k-means algorithm in terms of both accuracy and purity on the seven datasets. 

\begin{figure}[htbp]
\centering    
 
\subfigure[] 
{
	\begin{minipage}[t]{0.28\textwidth}
	\centering          
	\includegraphics[scale=0.26]{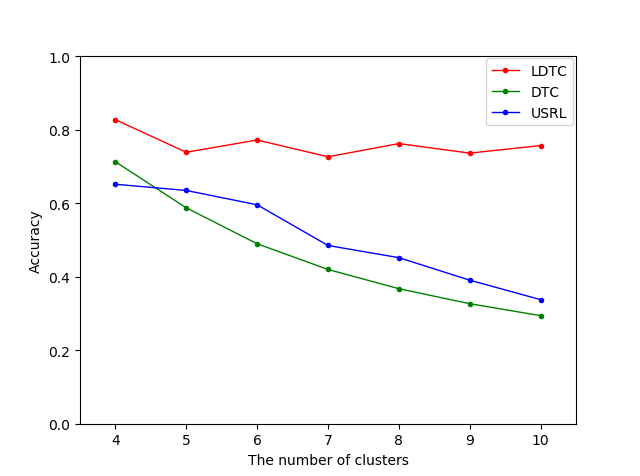}   
	\end{minipage}
}
\subfigure[] 
{
	\begin{minipage}[t]{0.28\textwidth}
	\centering      
	\includegraphics[scale=0.26]{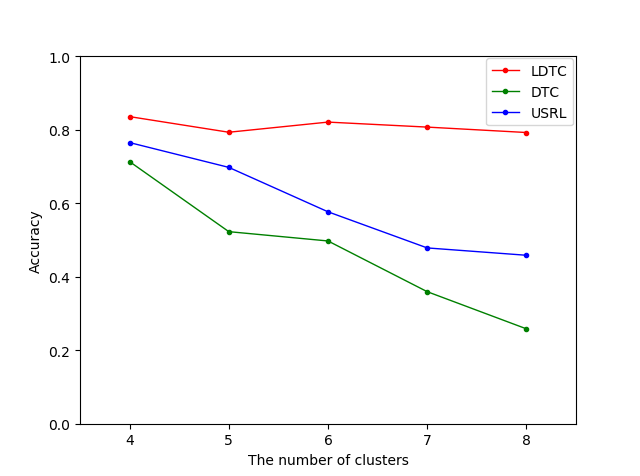}   
	\end{minipage}
}
\subfigure[] 
{
	\begin{minipage}[t]{0.28\textwidth}
	\centering      
	\includegraphics[scale=0.26]{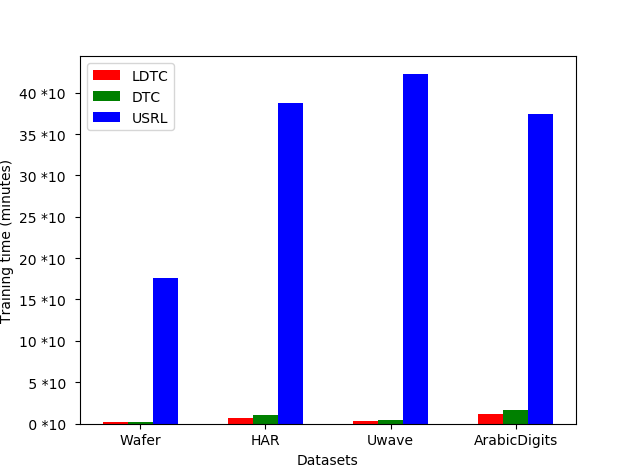}   
	\end{minipage}
}
\caption{The algorithm performance for learning new tasks. (a) The performance on dataset ArabicDigits. (b) The performance on dataset Uwave. (c) The training time of models: LDTC, DTC and USRL, respectively.} 
\label{fig2}  
\end{figure}

2) The proposed LDTC has a superior performance on different datasets than that of all the baselines. It is particularly worth mentioning that even though the USRL adopts a deeper network structure and uses an over 15 times longer train time than our proposed LDTC (see Figure \ref{fig2}(c)), the USRL's accuracy and purity are still lower than that of our LDTC. 

3) The LDTC is particularly good at clustering the longer, multi-variables and multi-category time series, for example, on datasets NetFlow (N=1337, \textbf{L}=994, \textbf{V}=4, C=2) and  ArabicDigits (N=8799, L=93, \textbf{V}=13, \textbf{C}=10), the performance of the LDTC is on average 17\% (Accuracy)/16\% (Purity)  higher than that of the other deep baselines DTC and USRL, up to 21\%.

The above experimental results clearly show that the LDTC's deep network model design, the multi-objective optimization strategy, and the deep temporal clustering method for multivariate time series are all effective and efficient.

\textbf{Lifelong temporal clustering.} One desired outcome of our LDTC has the ability to learn new tasks from non-stationary input data without catastrophic forgetting or degeneration of the model performance compared with other baselines. To this end, we investigated a number of different evaluation settings: i.i.d., where the model sees shuffled training data; sequential, where the model sees classes/clusters sequentially; and continuous drift, similar to the sequential case, but with classes gradually introduced by slowly increasing the number of samples from the new class within a batch.

Due to space limitations, here we only report the experiments (the average of ten experiments) on the two typical datasets ArabicDigits and Uwave. It is noted that some trained models have been stored in our LDTC's Model Pool. For the dataset ArabicDigits for LDTC's dynamic model expansion, there are trained models with the number of clusters of 6 and 8, respectively, and for the dataset Uwave, there is the trained model for the clusters 6. The experiments are shown in Figure \ref{fig2}.

Since the baselines are unable to learn new tasks, i.e., without the ability of lifelong temporal clustering learning, they are all in catastrophic forgetting or degeneration of the model performance when encountering new tasks contrary to our LDTC. Moreover, when there are some trained models in its Model Pool, our LDTC would gain better performances due to the fact that LDTC will make the trained model further refined in the input new tasks' i.i.d. settings.

\section{Conclusion}

This paper presents a novel lifelong deep temporal  clustering multivariate time series algorithm, or LDTC. It can effectively integrate dimensionality reduction and temporal clustering into an end-to-end deep unsupervised learning framework. With a specifically designed autoencoder CTAE and joint optimization in a new hierarchical way to both the latent representation learning and clustering objective, proposed LDTC achieves high-quality clustering results. Moreover, unlike all the previous works, LDTC has the lifelong temporal clustering unique ability, with which it can learn new concepts/classes and tackle the dynamic data during the unsupervised sequential learning tasks without the catastrophic forgetting or degradation of the model accuracy. Experiments on seven real-world multivariate time series datasets with different data distributions, sample size and clusters show that the LDTC is a promising method for dealing with temporal clustering issues effectively and efficiently.

Although the LDTC surpasses existing algorithms in terms of the clustering accuracy and the ability of lifelong temporal clustering learning, its clustering accuracy needs to be further improved. In addition, its lifelong learning mechanism embedded may also lead to the memory explosion. The more optimized solutions for these issues are our future research concern.

\section{Acknowledgments}
This work of Z. Wang, Y. Li, P. Zheng and Y. Jiao was supported by the National Natural Science Foundation of China (No.~62176202).







\small
\bibliographystyle{unsrt}
\bibliography{mybib}

@article{aghabozorgi2015time,
  title={Time-series clustering--a decade review},
  author={Aghabozorgi, Saeed and Shirkhorshidi, Ali Seyed and Wah, Teh Ying},
  journal={Information Systems},
  volume={53},
  pages={16--38},
  year={2015},
  publisher={Elsevier}
}

@article{yi2017grouped,
  title={Grouped convolutional neural networks for multivariate time series},
  author={Yi, Subin and Ju, Janghoon and Yoon, Man-Ki and Choi, Jaesik},
  journal={arXiv preprint arXiv:1703.09938},
  year={2017}
}

@book{brockwell1991time,
  title={Time series: theory and methods: theory and methods},
  author={Brockwell, Peter J and Davis, Richard A and Fienberg, Stephen E},
  year={1991},
  publisher={Springer Science \& Business Media}
}

@article{madiraju2018deep,
  title={Deep temporal clustering: Fully unsupervised learning of time-domain features},
  author={Madiraju, Naveen Sai and Sadat, Seid M and Fisher, Dimitry and Karimabadi, Homa},
  journal={arXiv preprint arXiv:1802.01059},
  year={2018}
}

@inproceedings{franceschi2019unsupervised,
  title={Unsupervised scalable representation learning for multivariate time series},
  author={Franceschi, Jean-Yves and Dieuleveut, Aymeric and Jaggi, Martin},
  booktitle={Advances in Neural Information Processing Systems},
  pages={4652--4663},
  year={2019}
}

@inproceedings{zheng2014time,
  title={Time series classification using multi-channels deep convolutional neural networks},
  author={Zheng, Yi and Liu, Qi and Chen, Enhong and Ge, Yong and Zhao, J Leon},
  booktitle={International Conference on Web-Age Information Management},
  pages={298--310},
  year={2014},
  organization={Springer}
}

@article{pei2017multivariate,
  title={Multivariate time-series classification using the hidden-unit logistic model},
  author={Pei, Wenjie and Dibeklio{\u{g}}lu, Hamdi and Tax, David MJ and van der Maaten, Laurens},
  journal={IEEE transactions on neural networks and learning systems},
  volume={29},
  number={4},
  pages={920--931},
  year={2017},
  publisher={IEEE}
}

@article{karim2019multivariate,
  title={Multivariate lstm-fcns for time series classification},
  author={Karim, Fazle and Majumdar, Somshubra and Darabi, Houshang and Harford, Samuel},
  journal={Neural Networks},
  volume={116},
  pages={237--245},
  year={2019},
  publisher={Elsevier}
}

@inproceedings{paneri2019regularizing,
  title={Regularizing fully convolutional networks for time series classification by decorrelating filters},
  author={Paneri, Kaushal and Vishnu, TV and Malhotra, Pankaj and Vig, Lovekesh and Shroff, Gautam},
  booktitle={Proceedings of the AAAI Conference on Artificial Intelligence},
  volume={33},
  pages={10003--10004},
  year={2019}
}

@article{oord2016wavenet,
  title={Wavenet: A generative model for raw audio},
  author={Oord, Aaron van den and Dieleman, Sander and Zen, Heiga and Simonyan, Karen and Vinyals, Oriol and Graves, Alex and Kalchbrenner, Nal and Senior, Andrew and Kavukcuoglu, Koray},
  journal={arXiv preprint arXiv:1609.03499},
  year={2016}
}

@article{bai2018empirical,
  title={An empirical evaluation of generic convolutional and recurrent networks for sequence modeling},
  author={Bai, Shaojie and Kolter, J Zico and Koltun, Vladlen},
  journal={arXiv preprint arXiv:1803.01271},
  year={2018}
}

@article{hochreiter1997long,
  title={Long short-term memory},
  author={Hochreiter, Sepp and Schmidhuber, J{\"u}rgen},
  journal={Neural computation},
  volume={9},
  number={8},
  pages={1735--1780},
  year={1997},
  publisher={MIT Press}
}

@inproceedings{graves2005bidirectional,
  title={Bidirectional LSTM networks for improved phoneme classification and recognition},
  author={Graves, Alex and Fern{\'a}ndez, Santiago and Schmidhuber, J{\"u}rgen},
  booktitle={International Conference on Artificial Neural Networks},
  pages={799--804},
  year={2005},
  organization={Springer}
}

@article{olah2015understanding,
  title={Understanding lstm networks},
  author={Olah, Christopher},
  year={2015}
}

@inproceedings{chen2017recurrent,
  title={Recurrent attention network on memory for aspect sentiment analysis},
  author={Chen, Peng and Sun, Zhongqian and Bing, Lidong and Yang, Wei},
  booktitle={Proceedings of the 2017 conference on empirical methods in natural language processing},
  pages={452--461},
  year={2017}
}

@inproceedings{fu2019dual,
  title={Dual attention network for scene segmentation},
  author={Fu, Jun and Liu, Jing and Tian, Haijie and Li, Yong and Bao, Yongjun and Fang, Zhiwei and Lu, Hanqing},
  booktitle={Proceedings of the IEEE Conference on Computer Vision and Pattern Recognition},
  pages={3146--3154},
  year={2019}
}

@incollection{mccloskey1989catastrophic,
  title={Catastrophic interference in connectionist networks: The sequential learning problem},
  author={McCloskey, Michael and Cohen, Neal J},
  booktitle={Psychology of learning and motivation},
  volume={24},
  pages={109--165},
  year={1989},
  publisher={Elsevier}
}

@article{goodfellow2013empirical,
  title={An empirical investigation of catastrophic forgetting in gradient-based neural networks},
  author={Goodfellow, Ian J and Mirza, Mehdi and Xiao, Da and Courville, Aaron and Bengio, Yoshua},
  journal={arXiv preprint arXiv:1312.6211},
  year={2013}
}

@inproceedings{xie2016unsupervised,
  title={Unsupervised deep embedding for clustering analysis},
  author={Xie, Junyuan and Girshick, Ross and Farhadi, Ali},
  booktitle={International conference on machine learning},
  pages={478--487},
  year={2016}
}

@misc{richard1961adaptive,
  title={Adaptive control processes: A guided tour},
  author={Richard, Bellman},
  year={1961},
  publisher={Princeton University Press}
}

@inproceedings{donahue2015long,
  title={Long-term recurrent convolutional networks for visual recognition and description},
  author={Donahue, Jeffrey and Anne Hendricks, Lisa and Guadarrama, Sergio and Rohrbach, Marcus and Venugopalan, Subhashini and Saenko, Kate and Darrell, Trevor},
  booktitle={Proceedings of the IEEE conference on computer vision and pattern recognition},
  pages={2625--2634},
  year={2015}
}

@inproceedings{zeiler2011adaptive,
  title={Adaptive deconvolutional networks for mid and high level feature learning},
  author={Zeiler, Matthew D and Taylor, Graham W and Fergus, Rob},
  booktitle={2011 International Conference on Computer Vision},
  pages={2018--2025},
  year={2011},
  organization={IEEE}
}

@article{maaten2008visualizing,
  title={Visualizing data using t-SNE},
  author={Maaten, Laurens van der and Hinton, Geoffrey},
  journal={Journal of machine learning research},
  volume={9},
  number={Nov},
  pages={2579--2605},
  year={2008}
}

@article{kingma2014adam,
  title={Adam: A method for stochastic optimization},
  author={Kingma, Diederik P and Ba, Jimmy},
  journal={arXiv preprint arXiv:1412.6980},
  year={2014}
}

@article{parisi2019continual,
  title={Continual lifelong learning with neural networks: A review},
  author={Parisi, German I and Kemker, Ronald and Part, Jose L and Kanan, Christopher and Wermter, Stefan},
  journal={Neural Networks},
  year={2019},
  publisher={Elsevier}
}

@inproceedings{hu2019overcoming,
  title={Overcoming catastrophic forgetting for continual learning via model adaptation},
  author={Hu, Wenpeng and Lin, Zhou and Liu, Bing and Tao, Chongyang and Tao, Zhengwei Tao and Zhao, Dongyan and Ma, Jinwen and Yan, Rui},
  booktitle={International Conference on Learning Representations},
  year={2019}
}

@inproceedings{rao2019continual,
  title={Continual Unsupervised Representation Learning},
  author={Rao, Dushyant and Visin, Francesco and Rusu, Andrei and Pascanu, Razvan and Teh, Yee Whye and Hadsell, Raia},
  booktitle={Advances in Neural Information Processing Systems},
  pages={7645--7655},
  year={2019}
}

@article{smith2019unsupervised,
  title={Unsupervised continual learning and self-taught associative memory hierarchies},
  author={Smith, James and Baer, Seth and Kira, Zsolt and Dovrolis, Constantine},
  journal={arXiv preprint arXiv:1904.02021},
  year={2019}
}

@article{munoz2019unsupervised,
  title={Unsupervised learning to overcome catastrophic forgetting in neural networks},
  author={Mu{\~n}oz-Mart{\'\i}n, Irene and Bianchi, Stefano and Pedretti, Giacomo and Melnic, Octavian and Ambrogio, Stefano and Ielmini, Daniele},
  journal={IEEE Journal on Exploratory Solid-State Computational Devices and Circuits},
  volume={5},
  number={1},
  pages={58--66},
  year={2019},
  publisher={IEEE}
}

@article{schafer2017multivariate,
  title={Multivariate time series classification with WEASEL+ MUSE},
  author={Sch{\"a}fer, Patrick and Leser, Ulf},
  journal={arXiv preprint arXiv:1711.11343},
  year={2017}
}

@misc{lichman2013uci,
  title={UCI machine learning repository},
  author={Lichman, Moshe and others},
  year={2013},
  publisher={Irvine, CA}
}

@inproceedings{shin2017continual,
  title={Continual learning with deep generative replay},
  author={Shin, Hanul and Lee, Jung Kwon and Kim, Jaehong and Kim, Jiwon},
  booktitle={Advances in Neural Information Processing Systems},
  pages={2990--2999},
  year={2017}
}

@inproceedings{kmeans1967,
  title={Some methods for classification and analysis of multivariate observations},
  author={MacQueen, James and others},
  booktitle={Proceedings of the fifth Berkeley symposium on mathematical statistics and probability},
  volume={1},
  number={14},
  pages={281--297},
  year={1967},
  organization={Oakland, CA, USA}
}

@inproceedings{nigam2000analyzing,
  title={Analyzing the effectiveness and applicability of co-training},
  author={Nigam, Kamal and Ghani, Rayid},
  booktitle={Proceedings of the ninth international conference on Information and knowledge management},
  pages={86--93},
  year={2000}
}

@article{chen2018lifelong,
  title={Lifelong machine learning},
  author={Chen, Zhiyuan and Liu, Bing},
  journal={Synthesis Lectures on Artificial Intelligence and Machine Learning},
  volume={12},
  number={3},
  pages={1--207},
  year={2018},
  publisher={Morgan \& Claypool Publishers}
}

@article{pedregosa2011scikit,
  title={Scikit-learn: Machine learning in Python},
  author={Pedregosa, Fabian and Varoquaux, Ga{\"e}l and Gramfort, Alexandre and Michel, Vincent and Thirion, Bertrand and Grisel, Olivier and Blondel, Mathieu and Prettenhofer, Peter and Weiss, Ron and Dubourg, Vincent and others},
  journal={the Journal of machine Learning research},
  volume={12},
  pages={2825--2830},
  year={2011},
  publisher={JMLR. org}
}

@inproceedings{raina2007self,
  title={Self-taught learning: transfer learning from unlabeled data},
  author={Raina, Rajat and Battle, Alexis and Lee, Honglak and Packer, Benjamin and Ng, Andrew Y},
  booktitle={Proceedings of the 24th international conference on Machine learning},
  pages={759--766},
  year={2007}
}

\end{document}